\documentclass{article}
\usepackage{spconf,amsmath,graphicx}
\usepackage{amsfonts} 
\usepackage{hyperref}

\usepackage{acro}
\usepackage{tabularx}
\usepackage{algorithm}
\usepackage{algpseudocode}
\usepackage{amsmath,amssymb,amsfonts}
\usepackage{multirow}

\usepackage{graphicx}
\usepackage{textcomp}
\usepackage{xcolor}

\usepackage[caption=false]{subfig}
\usepackage{xspace}
\usepackage{booktabs}

\title{Data-Efficient Stream-Based Active Distillation For Scalable Edge Model Deployment}

\name{\added[id=DM]{Dani Manjah*, Tim Bary*, Benoît Gérin, Benoît Macq and Christophe de Vleeschouwer\thanks{*D.~Manjah and T.~Bary have equal contributions.}\thanks{D.~Manjah, T.~Bary and B.~Gérin are funded by the Walloon region under grant No. 8881 (PIT ATMP) and No. 2010235 (ARIAC).}}}
\address{\added[id=DM]{ICTEAM, UCLouvain, 1348 Louvain-la-Neuve, Belgium}}

\newcommand{\eg}{\textit{e.g.}\xspace}
\newcommand{\ie}{\textit{i.e.}\xspace}

\usepackage[final]{changes}      
\definechangesauthor[name={Dani}, color=green]{DM}  

\DeclareAcronym{DNN}{
  short={DNN},
  long={Deep Neural Network}
}

\DeclareAcronym{CNN}{
  short={CNN},
  long={Convolutional Neural Network}
}

\DeclareAcronym{ML}{
  short={ML},
  long={Machine Learning}
}

\DeclareAcronym{WALT}{
  short={WALT},
  long={Watch and Learn Time-lapse}
}

\DeclareAcronym{TTA}{
  short={TTA},
  long={Test-Time Adaptation}
}

\DeclareAcronym{HoL}{
  short={HoL},
  long={Holonic Learning}
}

\DeclareAcronym{MAS}{
  short={MAS},
  long={Multi-Agent System}
}

\DeclareAcronym{HMAS}{
  short={HMAS},
  long={Holonic Multi-Agent System}
}

\DeclareAcronym{OMAS}{
  short={OMAS},
  long={Organizational Multi-Agent System}
}

\DeclareAcronym{CCTV}{
  short={CCTV},
  long={Closed-Circuit Television}
}

\DeclareAcronym{KD}{
  short={KD},
  long={Knowledge Distillation}
}

\DeclareAcronym{PBAL}{
  short={PBAL},
  long={Pool-Based Active Learning}
}

\DeclareAcronym{CL}{
  short={CL},
  long={Continual Learning}
}

\DeclareAcronym{SBAL}{
  short={SBAL},
  long={Stream-Based Active Learning}
}

\DeclareAcronym{AL}{
  short={AL},
  long={Active Learning}
}

\DeclareAcronym{SBAD}{
  short={SBAD},
  long={Stream-Based Active Distillation}
}

\DeclareAcronym{AD}{
  short={AD},
  long={Active Distillation}
}

\DeclareAcronym{CSBAD}{
  short={CSBAD},
  long={Clustered Stream-Based Active Distillation}
}

\DeclareAcronym{UDA}{
  short={UDA},
  long={Unsupervised Domain Adaptation}
}

\DeclareAcronym{MC}{
  short={MC},
  long={Model Compression}
}

\begin{document}
\maketitle
\begin{abstract}
Edge camera-based systems are continuously expanding, facing ever-evolving environments that require regular model updates. 
In practice, complex teacher models are run on a central server to annotate data, which is then used to train smaller models tailored to the edge devices with limited computational power. This work explores how to select the most useful images for training to maximize model quality while keeping transmission costs low. Our work shows that, for a similar training load (\ie, iterations), a high-confidence stream-based strategy coupled with a diversity-based approach produces a high-quality model with minimal dataset queries.

\end{abstract}

\begin{keywords}
Knowledge distillation, active distillation, dataset pruning, edge computing.
\end{keywords}    
\section{Introduction}
\label{sec:intro}

Edge camera systems are a cost-effective and privacy-preserv-ing solution for visual analytics tasks such as object detection and classification. As these networks scale to improve coverage and resilience~\cite{Perera2014}, each added sensor introduces a new visual domain, making it infeasible to rely on a single, universal \ac{DNN}. Additionally, existing sensors are subject to distribution shifts over time~\cite{CSBAD}, requiring frequent model updates.

However, the limited compute and memory of edge devices constrain \ac{DNN} size and performance~\cite{limitedRepresentation,yolov8}. Scalable deployment pipelines are \deleted{therefore} essential \replaced{for reducing}{to reduce} retraining costs and maintaining system reliability.

\ac{SBAD} addresses these challenges by training lightweight \emph{Student} models locally, using data pseudo-labeled by a powerful \emph{Teacher} model~\cite{sbad}. Its on-the-fly confidence-based sampling selects only images the Student is most confident about, reducing noisy labels and enabling continuous adaptation.

\begin{figure}[t]
    \centering
    \includegraphics[width=\linewidth]{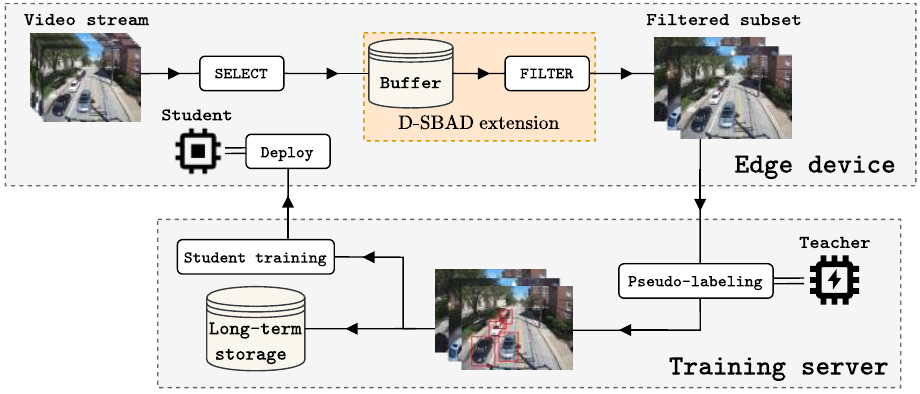}
    \caption{\deleted[id=DM]{Illustration of the proposed edge device model fine-tuning pipeline.}Candidate frames extracted from the device stream are stored in a buffer, where a pool-based filtering module \deleted{(D-SBAD extension)} selects high-quality samples to reduce server-bound training data while maintaining dataset quality. The selected frames are then pseudo-annotated by a teacher model to train the next iteration of the edge model.}
    \label{fig:DSBAD}
\end{figure}

While effective, confidence alone can result in redundant datasets and unnecessary data transfer. To improve efficiency, we introduce D-\ac{SBAD}, which adds an on-device filtering stage to \ac{SBAD}. As shown in Fig.~\ref{fig:DSBAD}, candidate frames are buffered and filtered using a lightweight embedding model that selects a diverse subset of samples for training.

Our experiments on 15 WALT cameras~\cite{Reddy_2022_CVPR} show that this two-stage selection significantly reduces bandwidth and training costs while preserving model performance. By maximizing embedding diversity with minimal overhead, D-\ac{SBAD} enhances the scalability of edge learning systems.

\deleted[id=DM]{The remainder of the paper is organized as follows: Section~\ref{sec:SOTA} reviews related work, Section~\ref{sec:samplesselection} details the D-\ac{SBAD} framework, Section~\ref{sec:materialsSBAD} describes our experimental setup, and Section~\ref{sec:filterResults} presents evaluation results.}

\section{Related Work}
\label{sec:SOTA}
\deleted[id=DM]{
This section summarizes prior work on knowledge distillation, active learning, and dataset efficiency, providing the foundation for our proposed framework.}
\subsection{Online Knowledge Distillation}
\label{sec:modelCompression}
\ac{KD} encompasses techniques where a \emph{Student} model learns from another \emph{Teacher} model~\cite{hinton2015distilling}. This approach can be used to create compact models that are based on the capabilities of a more complex model, and is especially useful when deploying on edge devices, where storage and computational capacity become critical factors~\cite{cheng2017survey}. Through regular updates, the \emph{Teacher}'s knowledge ensures that the \emph{Student} model continues to perform effectively even when operating on changing data distributions thanks to its greater robustness~\cite{saito2019semi}.  

Recently, \acs{KD} has supported innovative applications in video analytics, particularly through \textit{Online Distillation}~\cite{CSBAD,boldoQuery}, where a compact model's weights are updated in real-time to mimic the output of a larger, pre-existing model. \deleted[id=DM]{Other notable applications include \textit{Context- and Group-Aware Distillation} \cite{rivas2022towards,delta_distillation_2022}, which accounts for context and/or sub-population shifts when delivering \emph{Student} models.}
\subsection{Active Distillation}
\label{sec:activeLearning}
Typical \textbf{\acl{AL}} assumes the presence of a perfect annotator, an assumption that does not hold in a distillation framework where pseudo-labeling can lead to the accumulation of errors due to incorrect predictions (\ie, \emph{confirmation bias}) in semi-supervised or unsupervised learning~\cite{arazo2020pseudo}. To mitigate noisy labels in \replaced[id=DM]{KD}{a knowledge distillation scheme}, the \textbf{\ac{AD}} framework has been proposed by~\cite{aid,sbad,Gerin_2024_CVPR} as a query-efficient \ac{AL} method that is robust to imperfections in the \emph{Teacher} model. Specifically, in \textbf{\ac{SBAD}}~\cite{sbad}, the authors demonstrated that high-confidence automatic labeling of training samples plays a crucial role in the \ac{AD} selection process. By selecting samples where the Student model exhibits high confidence, the framework reduces labeling errors and enhances the reliability of the distillation process. In this work, we retain the \textbf{\ac{SBAD}} and its top-confidence selection rule due to its effectiveness in handling large volumes of unlabeled images and its ability to continuously adapt to incoming cameras without requiring labeled data at test time~\cite{testTimeAdaptation}. However, we propose a two-stage approach that balances exploration and exploitation, similar to \textbf{PPAL}~\cite{Yang2022}. In that sense, we extend \ac{SBAD} \textbf{with a filtering module designed to identify a subset that is representative of the camera's domain after exploration.} 

\subsection{Pool-Based Active Learning}
\label{sec:poolbased}
\deleted[id=DM]{Our second stage pipeline aims to solve the problem of selecting an informative subset from the pool of data available in the buffer.} Pool-based methods \replaced[id=DM]{aim to select an informative subset from the pool of data available in the buffer.}{can be broadly classified as follows:}  
\textbf{Uncertainty-based} methods select samples based on predictive confidence scores or error rates. The \textbf{Minmax} strategy~\cite{RoyUnmesh} selects the least confident samples from the unlabeled pool, assuming that low-confidence predictions are more informative for training.
\deleted{-}\textbf{Error-based} strategies refine this process by selecting samples where predicted and actual errors diverge. Wu et al.~\cite{WuEntropy} proposed \textbf{Entropy-Diversity Sampling}, combining entropy-based NMS (Non-Maximum Suppression) with a diversity-based prototype selection phase to maximize intra- and inter-class variation.  
\deleted{-}\textbf{Density-based} methods select samples that best represent the underlying data distribution. Yang et al.~\cite{Yang2022} proposed \textbf{PPAL}, a two-stage strategy using uncertainty sampling followed by category-conditioned similarity matching to ensure diversity in the selected set.

\subsection{Dataset Pruning}
\label{sec:datasetEffeciency}
Dataset pruning belongs to coreset techniques with the focus of selecting a representative subset from a larger dataset to maintain or enhance model performance while reducing training data size~\cite{feldman2020turning}. \textbf{Entropy}~\cite{coleman2020dataEntropy} is a classic approach but primarily limited to classification tasks. Many state-of-the-art methods, such as \textbf{CSOD}~\cite{CSOD} or \textbf{AUM}~\cite{AUM}, rely on ground truth labels. \textbf{CSOD} trains a model to identify informative samples, while \textbf{AUM} removes mislabeled or ambiguous data using ground truth information. 
In edge computing environments, utilizing the edge model in a self-supervised manner can introduce confirmation bias, particularly given the model's limited capacity. Alternatively, querying a larger, general-purpose \emph{Teacher} model, or an \emph{Oracle}, conflicts with scalability objectives due to increased bandwidth usage and annotation costs associated with transmitting images. 

Self-supervised methods that compute per-sample importance scores, such as distance to other samples~\cite{zhang2020deepemd}, or that utilize clustering heuristics to identify representative samples~\cite{Yang2022}, offer alternatives. Finally, \textbf{MODERATE}~\cite{Moderate} employs redundancy and representativeness strategies for dynamic and moderate pruning, while \textbf{TFDP} introduces a training-free dataset pruning approach for instance segmentation tasks~\cite{dai2025trainingfreedatasetpruninginstance}. In this work, we benchmark \textbf{MODERATE} and \textbf{TFDP} as they focus on reducing redundancy and selecting representative samples without any labels\replaced[id=DM]{.}{, are more suitable in these cases.} 

\section{D-SBAD}\label{sec:samplesselection}
Our framework follows a client-server dynamic, where a lightweight, pre-trained \emph{Student} model \( \theta \), deployed on an edge device (\eg, a camera), processes a video stream \( \mathbf{X} \). To ensure the \emph{Student} remains up-to-date, it is periodically fine-tuned on a server using a curated set $\mathcal{T}$ of training samples selected from $\mathbf{X}$. The server hosts a larger, general-purpose \emph{Teacher} model \( \Theta \), which generates pseudo-labels for the selected images. These pseudo-labels act as ground truth during fine-tuning of $\theta$. Since transmitting and storing all frames on the server is infeasible in practice, the selected set $\mathcal{F}$ must adhere to a frame budget~$B$. Our objective is therefore to select the $B$ frames from $\mathbf{X}$ that maximize the \emph{Student}'s performance after training.

\subsection{Sample Selection}
Due to the client's limited storage, it cannot retain the entire video stream for retrospective sampling. Instead, the client must decide in real time whether an image $I_t$ is informative enough to be transmitted to the server. The baseline approach from \cite{sbad} selects and immediately transmits $B$ images to the remote server using a given \texttt{SELECT} strategy. We propose an enhanced selection pipeline incorporating a filtering step. Instead of selecting exactly $B$ images upfront, the client first applies \texttt{SELECT} to construct for an acquisition period $T$, an initial ``candidate set" $\mathcal{S}$  which is temporarily stored in an image buffer. This set is then refined \deleted[id=DM]{using a pool-based filtering strategy, \texttt{FILTER},} to obtain the final subset $\mathcal{F}$ of size $B$. This additional step improves the likelihood of selecting high-quality images while maintaining the same bandwidth and storage constraints on the server side. The modified pipeline is illustrated in Figure~\ref{fig:DSBAD} and detailed in Algorithm~\ref{alg:2SBAD}.

\begin{algorithm}
\caption{D-SBAD}\label{alg:2SBAD}
\begin{algorithmic}[1]
\Require \emph{Student} model $\theta$, \emph{Teacher} model $\Theta$, training frame budget $B \geq 1$, video stream \(\mathbf{X}\), window of collection $T$, \texttt{SELECT} strategy,  \texttt{FILTER} strategy.
\Ensure $\theta$, a fine-tuned \emph{Student} model.
\State $\mathcal{S} \gets \emptyset $ 
\State $t \gets 0$ \Comment{Timestamp}
\For{$t = 1,\dots, T$}
    \State Observe current frame $I_t$ from $\mathbf{X}$
        \If {\texttt{SELECT}($I_t$) $=$ TRUE}
          \State $\mathcal{S} \gets \mathcal{S} \cup (I_t, \; \text{Labels}=\emptyset)$
        \EndIf
    \EndFor
    \State $\mathcal{F} \gets \texttt{FILTER}(\mathcal{S},B)$
    
\For{$ I_k \in \mathcal{\mathcal{F}}$}
    \State $\tilde{P}_k \gets \Theta(I_k)$ \Comment{Infer pseudo-labels}
    \State $\text{Labels}(I_k) \gets (\tilde{P}_k) $ \Comment{Update Pseudo-Labels}
\EndFor

\State $\theta \gets train(\theta, \mathcal{F})$

\end{algorithmic}
\end{algorithm}

\section{Materials and Methods}
\label{sec:materialsSBAD}

\deleted[id=DM]{This section outlines the dataset, models, training and sampling strategies as well as evaluation protocols used in our experiments. \textit{Note that the code to reproduce the experiments and the proposed framework will be publicly available upon acceptance.}}

\subsection{Dataset}
\label{sec:WALTdataset}
We use the \ac{WALT} dataset~\cite{Reddy_2022_CVPR}, which consists of footage from 15 cameras capturing vehicle circulation in public spaces over one to five weeks. The dataset includes various lighting, weather conditions, and viewpoints, with sample counts ranging from 5,000 to 40,000 frames per week due to varying recording rates. Each camera has a dedicated test set. Together, these test sets total 2450 images and 16,877 human-annotated vehicle instances, as per the instructions in~\cite{sbad}. Images are used without preprocessing, except for grouping COCO labels~\cite{COCO} ``bikes, cars, motorcycles, buses, and trucks" into a ``vehicle" category.

\subsection{Models}
\label{sec:StudenetModels}

We employ the YOLO11 architecture~\cite{yolov8} for both the student and teacher models: the \emph{Student} model $\theta$ is a YOLO11n (3.2M parameters), and the \emph{Teacher} $\Theta$ is a YOLO11x (68.2M parameters). Both are initialized with pre-trained weights from the COCO dataset~\cite{COCO}. The \emph{Student} model is fine-tuned with a constant iteration criterion. For a fixed batch size of 16, the number of epochs is adjusted in inverse proportion to the training set size $B$ to maintain the same computational cost across different set sizes. Other training parameters follow the defaults from~\cite{yolov8}.

\subsection{Evaluation}
The performance of the \emph{Student} models is evaluated using the mAP$_{50-95}$ metric, which calculates the mean Average Precision (mAP) at intersection-over-union (IoU) thresholds ranging from 0.50 to 0.95 in increments of 0.05.

\subsection{Initial Sampling} 

\replaced{We sample\deleted{d} using \texttt{TOP-CONFIDENCE}~\cite{sbad}, where an image is transmitted if the maximum \emph{Student's} confidence exceeds a threshold, computed as the $1 - \alpha$ quantile ($\alpha = 0.1$)  during a warm-up phase of size $w = 720$ images.}
{We employed the recommended \texttt{TOP-CONFIDENCE} strategy from~\cite{sbad} to generate an initial pool of candidates, which has been shown to provide clear and well-lit images with objects of interest~\cite{CSBAD}. The method relies on a threshold based on the highest confidence of the \emph{Student} model, computed as the $1 - \alpha$ quantile during a warm-up phase of size $w$. Each image is transmitted if the maximum confidence exceeds the threshold; otherwise, it is discarded. We set the parameters to $\alpha = 0.1$ and $w = 720$.}

\subsection{Filtering Strategies}
\label{sec:FILTERstrats}
We \deleted{provide a short description of} benchmark\deleted{ed} filtering methods, compatible with edge constraints \replaced[id=DM]{and that do not require ground-truth labels.}{well of ground truth labels.} 

\begin{itemize}
    \item \textbf{Farthest First (FF)}~\cite{geifman2017deep}: We provide a deterministic version of the FF algorithm. Namely, FF considers a function $\phi$ to map any image $i$ to its representation in a latent space. Given a selected set $\mathcal{F}$ and a candidate set $\mathcal{S}$, where $\mathcal{F} \cap \mathcal{S} = \emptyset$, the selection proceeds as follows:
\begin{enumerate}
    \item The first image $I_* \in \mathcal{S} $ is selected as the one lying in the densest region of the latent space:
\begin{equation}
    I_* = \arg \max_{I\in\mathcal{S}} \sum_{I'\in\mathcal{S}} \langle \phi(I), \phi(I') \rangle.
\end{equation}
    \item The remaining $B-1$ images are iteratively added such that they minimize  the maximum cosine similarity to the selected set:
\begin{equation}
    I_b = \arg \min_{I'\in\mathcal{S}} \max_{I\in\mathcal{F}} \text{cosim}(\phi(I), \phi(I')).
\end{equation}
\end{enumerate}
Our implementation follows Algorithm~\ref{alg:filter-diversity}. 

    \item \textbf{Training-Free Dataset Pruning (TFDP)}: Initially designed for image segmentation, TFDP~\cite{dai2025trainingfreedatasetpruninginstance} computes a \textit{Shape Complexity Score (SCS)} for each instance detected in an image. This score is given by:
    \begin{equation}
        S_{i,j} = \frac{P_{i,j}}{2\sqrt{\pi A_{i,j}}},
    \end{equation}
    where $i$ is the image, $j$ the instance, $P_{i,j}$ the instance mask perimeter, and $A_{i,j}$ the instance mask area. In our case, we employ the detection boxes as the segmentation mask. The $B$ images with the highest SCS sum constitute the final training set.

    \item \textbf{Moderate Coreset}~\cite{Moderate} first embeds all images $i$ into a latent space using a mapping function $\phi$. It then selects the $B$ images whose distances from the dataset center are closest to the median distance in the latent space.

\item \textbf{Least Confidence}~\cite{li2006confidence} selects the $B$ images from $\mathcal{S}$ for which the \emph{Student} model has the lowest confidence in its predicted labels.  

\item \textbf{Random} \replaced{selects}{is a standard baseline, selecting, where} images \replaced{with}{have} a uniform \replaced{probability}{chance of being selected}. We repeated the experiments with six seeds.  
\end{itemize}  


\begin{algorithm}[!t]
\caption{Deterministic Farthest First}
\label{alg:filter-diversity}
\begin{algorithmic}[1]
\Require Set of images $\mathcal{S}$, budget $B$, mapping function $\phi$.
\State $I_* \gets \arg \max_{I\in\mathcal{S}}\sum_{I'\in\mathcal{S}}\langle \phi(I), \phi(I')\rangle$
\State $\mathcal{F} \gets\{I_*\}$
\State $\mathcal{S} \gets \mathcal{S} \setminus\{I_*\}$
\For{$b\in[1,B-1]$}
    \State $I_b \gets \arg \min_{I'\in\mathcal{S}} \max_{I\in\mathcal{F}} \text{cosim} \left( \phi(I), \phi(I') \right)$
    \State $\mathcal{F} \gets \mathcal{F} \cup \{I_b\}$
    \State $\mathcal{S} \gets \mathcal{S} \setminus \{I_b\}$
\EndFor

\State \textbf{return} $\mathcal{F}$
\end{algorithmic}
\end{algorithm}

\section{Results and Discussion}
\label{sec:filterResults}
We evaluate filtering strategies, and the ratio between collected and transmitted image sets. The temporal collection period $T$ is experimentally simulated, such that the collected set size is a $\gamma$-multiple of the image budget B,~\ie,$\; |\mathcal{S}|=\gamma B$.

\subsection{Impact of Filtering Strategy}
\begin{figure}[htpb]
\centering
{
  \includegraphics[width=0.99\linewidth]{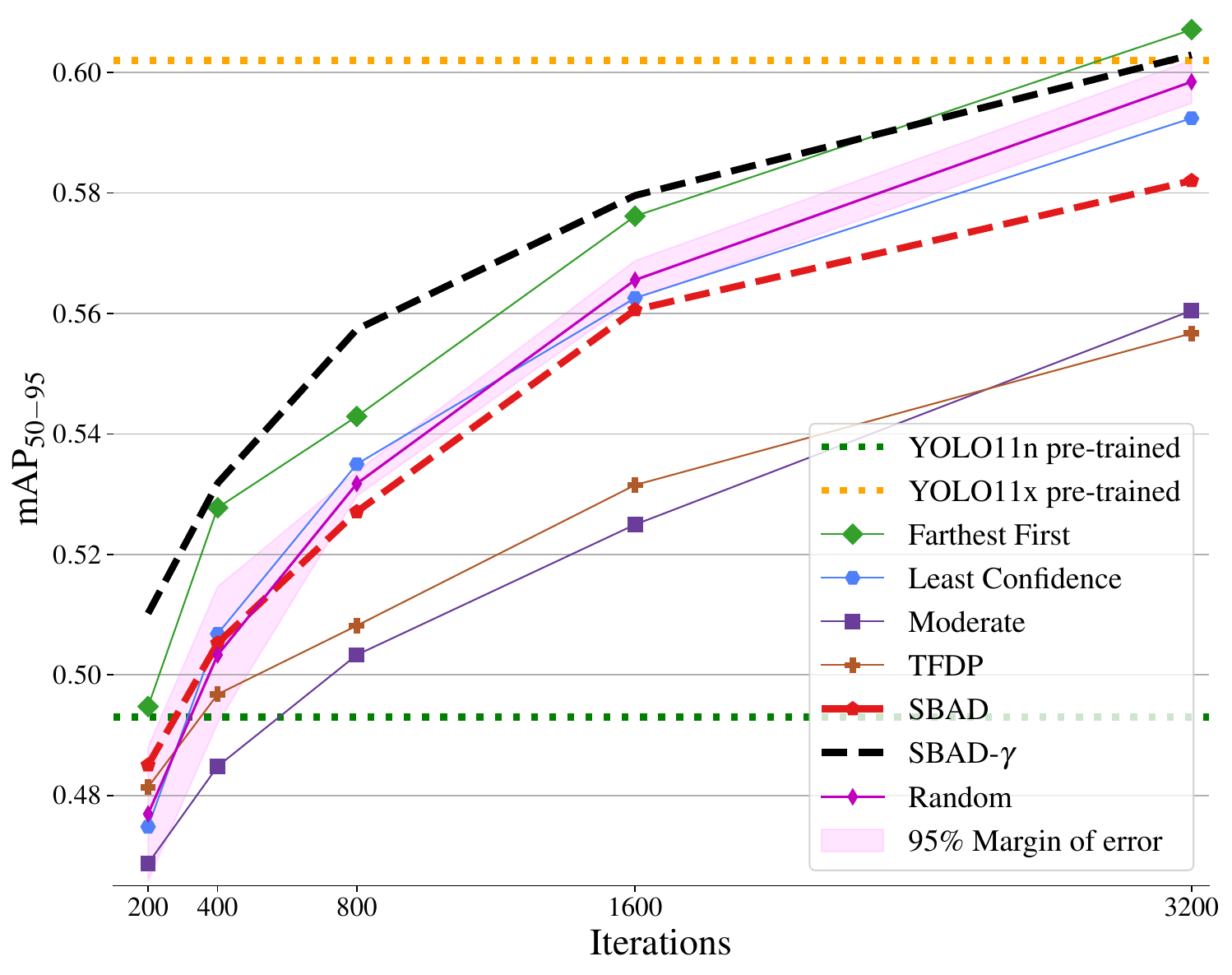}
  \caption{Performance comparison of \emph{Student} models trained on datasets from various \texttt{FILTER} strategies, with $\gamma=8$. All methods except SBAD-$\gamma$ use fixed 100 epochs. SBAD-$\gamma$ training epochs are adjusted by factor $1/\gamma$.}
  \label{fig:pruningImpact}
}
\end{figure}

Fig.~\ref{fig:pruningImpact} compares the average mAP$_{50-95}$ for filtering strategies described in Section~\ref{sec:FILTERstrats}, with parameters set as $\gamma=8$ and training for 100 epochs. Two baselines are included: SBAD selects $B$ frames without filtering ($|\mathcal{F}|=|\mathcal{S}|=B$), while SBAD-$\gamma$ uses the full collected set ($|\mathcal{F}|=|\mathcal{S}|=\gamma B$). Furthermore, we report the performance of pre-trained YOLO11 \emph{Student} and \emph{Teacher}. To ensure fair comparison, epochs for SBAD-$\gamma$ are reduced to match iteration counts. 

Among evaluated methods, \emph{Farthest First} (FF) yields the highest performance, comparable to SBAD-$\gamma$ despite using eight times fewer images. Conversely, both TFDP and Moderate Coreset perform below the unfiltered baseline, suggesting suboptimal selection strategies in this use-case. Furthermore, FF provides \emph{Students} superior to pretrained YOLO11n from 200 iterations and can outperform YOLO11x \emph{Teacher} at 3200 iterations. 
Future work could investigate alternative initial sampling schemes, \textit{e.g.,} the methods of \cite{boldoQuery,select} to broaden the range of starting points.


\subsection{Impact of the Candidate Set Size}
\label{sec:gammaImpact}
\begin{figure}[htpb]
    \centering
    \includegraphics[width=0.9\linewidth]{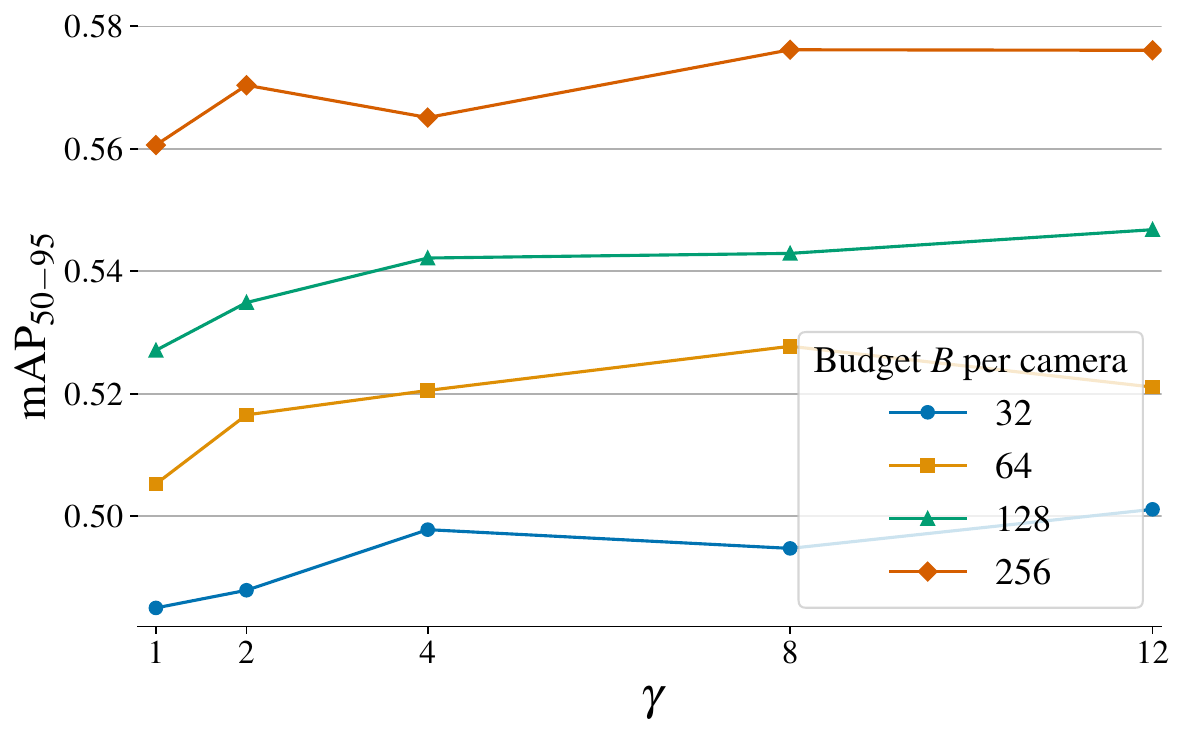}
    \caption{mAP$_{50-95}$ of the \emph{Student} models trained on datasets generated by the FF \texttt{FILTER} strategy, for different values of $\gamma$ and $B$. All models are trained for 100 epochs.}
    \label{fig:alphaStudy}
\end{figure}

The choice of the initial collection size ($\mathcal{S}$) impacts performance and client-side costs (buffer size and stream duration). Fig.~\ref{fig:alphaStudy} illustrates average mAP$_{50-95}$ for FF as a function of the temporal acquisition $T$, here defined by the multiplier $\gamma={1,2,4,8,12}$ and budget $B={32,64,128,256}$. 

Results highlight significant performance gains when increasing from no exploration ($\gamma=1$) to moderate exploration ($\gamma=2$). Further increases in $\gamma$ provide diminishing returns, suggesting an optimal trade-off at moderate collection set sizes.
To further enhance bandwidth performance, methods can rely on efficient and secure protocols~\cite{security,security2}.

\section{Conclusion}
We proposed a filtering step within the \ac{SBAD} framework to enhance scalability by reducing the amount of transmitted data while maintaining model performance. Experiments demonstrated that the Farthest First algorithm, optimizing latent space coverage, provided the best filtering performance, matching or exceeding baselines using substantially fewer images. Our results confirm the practicality and effectiveness of compact embeddings generated by edge-compatible ViT models such as DINOv2. Future work will investigate lightweight self-supervised pipelines that can be fully executed on edge devices, thereby eliminating the dependency on server-side. 

\bibliographystyle{IEEEbib}
\bibliography{main.bib}


\end{document}